\documentclass[journal]{IEEEtran}

\usepackage{xcolor,soul,framed} 
\usepackage[printonlyused]{acronym}
\usepackage{float}
\usepackage{subfig}
\usepackage{algorithm}
\usepackage{algpseudocode}
\usepackage{bbm}
\usepackage{comment}
\makeatletter
\newcommand\fs@norules{\def\@fs@cfont{\bfseries}\let\@fs@capt\floatc@ruled
  \def\@fs@pre{}%
  \def\@fs@post{}%
  \def\@fs@mid{\kern3pt}%
  \let\@fs@iftopcapt\iftrue}
\makeatother
\colorlet{shadecolor}{yellow}
\usepackage[pdftex]{graphicx}
\graphicspath{{../pdf/}{../jpeg/}}
\DeclareGraphicsExtensions{.pdf,.jpeg,.png}

\usepackage[cmex10]{amsmath}
\usepackage{array}
\usepackage{mdwmath}
\usepackage{mdwtab}
\usepackage{eqparbox}
\usepackage{url}
\usepackage{amsfonts}
\usepackage{cleveref}
\usepackage{subfig}
\usepackage[scr=rsfs]{mathalpha}
\usepackage{makecell}
\usepackage{float} 
\usepackage[utf8]{inputenc}
\usepackage{booktabs} 

\usepackage{mathtools}
\usepackage{array}
\newcolumntype{P}[1]{>{\centering\arraybackslash}p{#1}}
\newcolumntype{M}[1]{>{\centering\arraybackslash}m{#1}}

\newcommand{\bx}{\textbf{x}}

\newcommand{\bD}{\textbf{D}}
\newcommand{\bZ}{\textbf{Z}}

\newcommand{\bC}{\textbf{C}}
\newcommand{\bL}{\textbf{L}}

\newcommand{\bX}{\textbf{X}}

\newcommand{\be}{\textbf{e}}
\newcommand{\bz}{\textbf{z}}

\crefalias{subequation}{equation} 
\crefalias{eqnarray}{equation} 
\crefformat{pluraleq}{(#2#1#3)} 

\newcommand{\argmin}{\operatornamewithlimits{argmin}}
\usepackage{bm}

\newcommand{\rev}{\textcolor{black}}

\DeclareUnicodeCharacter{2212}{-}
\begin{document}
\bstctlcite{IEEEexample:BSTcontrol}
    \title{A generative model for surrogates of spatial-temporal wildfire nowcasting}
\author{Sibo Cheng, Yike Guo, and Rossella Arcucci
\thanks{corresponding: sibo.cheng.math@gmail.com}    

\thanks{Sibo Cheng, Yike Guo and  Rossella Arcucci are with Data Science Institute, Department of computing, Imperial College London, SW7 2AZ, UK. }
\thanks{Yike Guo is also with Department of Computer Science and Engineering, The Hong Kong University of Science and Technology , 999077, Hongkong, China. }
\thanks{Rossella Arcucci is also with Department of Earth Science and Engineering, Imperial College London, SW7 2AZ, UK. }
}

\markboth{accepted for publication in IEEE Transactions on Emerging Topics in Computational Intelligence,
2023}{Cheng {\textit{et al.}}: Spectral Cross-Domain Neural Network with Soft-adaptive Threshold Spectral
Enhancement}

\maketitle

\begin{abstract}
Recent increase in wildfires worldwide has led to the need for real-time fire nowcasting. Physics-driven models, such as cellular automata and computational fluid dynamics can provide high-fidelity fire spread simulations but they are computationally expensive and time-consuming. Much effort has been put into developing machine learning models for fire prediction. However, these models are often region-specific and require a substantial quantity of simulation data for training purpose. This results in a significant amount of computational effort for different ecoregions.
In this work, a generative model is proposed using a three-dimensional Vector-Quantized Variational Autoencoders to generate spatial-temporal sequences of unseen wildfire burned areas in a given ecoregion.
The model is tested in the ecoregion of a recent massive wildfire event in California, known as the Chimney fire. Numerical results show that the model succeed in generating coherent and structured fire scenarios, taking into account the impact from geophysical variables, such as vegetation and slope. Generated data are also used to train a surrogate model for predicting wildfire dissemination, which has been tested on both simulation data and the real Chimney fire event.
\end{abstract}


%
\IEEEpeerreviewmaketitle


\section{Introduction}\label{sec3}
A significant increase of wildfire frequency has been noticed world-widely in the past decades. These unwanted wildland fires can result in losses of lives, huge economic cost and lethal effects of air pollution. Real-time fire nowcasting is crucial for establishing fire fighting strategies and preventing multiple fire related hazards. Much effort has be given to develop fire simulation algorithms, for instance, based on \ac{CA}~\cite{ALEXANDRIDIS2008,freire2019using}, \ac{MTT}~\cite{finney2002fire} and \ac{CFD}~\cite{valero2021multifidelity}. These physics-driven modellings are usually built on complex (often probabilistic) mathematical models which take into account a considerable number of geophysical and climate features, such as vegetation distribution, slope elevation and wind directions.
As a consequence, running physics-based simulations for large fire events is computationally expensive and time consuming~\cite{papadopoulos2011comparative}. A key challenge in current wildfire forecasting study consists of developing efficient surrogate models (also known as digital twins)~\cite{zhu2022building,Cheng2022JCP,zhong2023reduced}, capable of near real-time fire nowcasting. 

In recent years, thanks to their effectiveness, machine learning techniques have been widely adapted to wildfire prediction problems~\cite{jain2020review}. For example, much attention is given in applying \acp{CNN} on terrestrial-based images for fire and smoke detection~\cite{saponara2021real,li2020image}. 
Shallow machine learning algorithms, such as \ac{RF} and \ac{SVM}, have been widely adopted in regional fire-susceptibility mapping~\cite{achu2021machine}. 
The very recent work of~\cite{Cheng2022JCP} proposed a deep learning surrogate model based on \ac{CAE} for reduced-order-modelling and \ac{RNN} for fire dynamic predictions. Their approach is tested in three recent large fire events in California where a \ac{CA} fire simulator is employed to generate the training dataset for the surrogate model. By learning from the simulated fire spread instances, the online computational time of fire spread forecasting can be significantly reduced from several hours to several seconds.
Furthermore, the model is capable of efficiently integrating near real-time observations from satellites via performing data assimilation in the reduced latent space. 
However, as stated in~\cite{Cheng2022JCP}, a main challenge for generalizing their model to different ecoregions stands for the generation of the training data using the \ac{CA} model. As mentioned in~\cite{Cheng2022JCP}, runing the CA model for simulating a massive wildfire event can take up to several days on a High performance computing (HPC) system. In this work, we aim to build a generative model that can produce spatial-temporal consistent and physical realistic wildfire scenarios in a given ecoregion.

Much effort has been given in generating high dimensional dynamical systems that are consistent with physical principles~\cite{pan2020physics,cheng2022spatio,otten2021event,cheng2020advanced,chagot2022surfactant}. Typical approaches include \acp{GAN}, Normalizing Flows and \acp{VAE}. Compared to the other methods, the advantages of \ac{VAE} are mainly two-folds: (i) \acp{VAE} are  Likelihood-based methods which are convenient to train and interpret~\cite{yan2021videogpt}; (ii) \acp{VAE} compress data into a low-dimensional latent space, enabling efficient interpolation between data points~\cite{chen2016infogan}. The latter is crucial for generating a smooth transition between different data samples. 
In particular, the Vector Quantized Variational Autoencoder (VQ-VAE) introduced by~\cite{van2017neural} is an extension of \ac{VAE} that replaces the continuous latent variables with discrete ones.
The main strength of  VQ-VAE consists of producing high-fidelity and sharp samples~\cite{yan2021videogpt}, which are vital for spatial-temporal systems. 
More precisely, by contructing a discrete, finite and trainable latent space (known as 'codebook'), VQ-VAE demonstrates exceptional capability in capturing the underlying structure of the data, enabling more coherent and structured generated samples~\cite{van2017neural}. Therefore, by implementing 3D convolutional layers (with an additional temporal axis), we apply \ac{VAE} for generating spatial-temporal fire scenarios in this study.

The proposed approach is tested in a recent massive wildfire event in California, known as the Chimney fire~\cite{kuligowski2020modelling}, which lasted more than 20 days and had burned out over $180 \textrm{km}^2$ of land. A \ac{CA} model~\cite{ALEXANDRIDIS2008} is first implemented with random ignition points in the ecoregion of the Chimney fire to generate training data. 
500 sequences of burned area evolution (each consists of 16 snapshots, equivalent to 4 days in real time) are generated using 3D VQ-VAE. Qualitative and quantitative analysis is performed to assess the fidelity of the generated samples in relation to established geophysical principles in wildfire spread~\cite{ALEXANDRIDIS2008}. Thanks to its great efficiency compared to the \ac{CA} model, the proposed generative model can provide a fast recognition of 'area at risk' for fire management.

A \ac{ML} surrogate model, based on \ac{POD} and \ac{RNN}, is also implemented in this study. \ac{POD} is a method used in the field of dynamical systems to decompose a complex system into a set of uncorrelated modes~\cite{chatterjee2000introduction}. These modes are chosen to represent the most important features of the dynamical system, and can be used to reconstruct the original system with a high degree of accuracy~\cite{xiao2015non}. On the other hand, much effort has been given in predicting compressed latent variables via machine learning models, such as \ac{RF}~\cite{gong2022data}, \ac{RNN}~\cite{amendola2020,Cheng2022JCP} or Transformers~\cite{geneva2022transformers}. Since we look for long-term predictions, \ac{LSTM} neural netowk~\cite{hochreiter1998vanishing}, a variant of \ac{RNN} is chosen to build the surrogate model in this study. 
\ac{LSTM} is a type of \ac{RNN} that is able to process  sequential data. \acp{LSTM} is able to maintain a "memory" of previous inputs, allowing them to better understand context and relationships in the data. The surrogate model implemented in this work uses the 500 fire spread sequences generated by VQ-VAE as training data. It significantly outperforms the surrogate model only trained using 40 \ac{CA} simulations on both synthetic fire events and the observed Chimney fire event. The observation of the burned area of the Chimney fire is processed using the MODIS satellite.
The proposed generative model runs xxx times faster than the \ac{CA} simulation, thus, it can  significantly release the computational burden of generating training samples for surrogate models. 

The main contribution of the present paper can be summarised as:
\begin{itemize}
    \item To the best of the author's knowledge, this is the first reported attempt of generative models applied on wildfire spread dynamics.
    \item The proposed approach is substantially faster than existing physics-based approaches, namely \ac{CA} and \ac{MTT}, in generating wildfire spread scenarios. 
    \item Generated wildfire data are used to train a machine learning surrogate model for fire spread prediction. The prediction accuracy on both the unseen fire simulation data and the real wildfire event can be significantly improved thanks to the generated data.
\end{itemize}

The rest of this paper is organised as follows. Section~\ref{sec:fire simulation} introduces the data source, the study area and the \ac{CA} fire spread model used in this study. Section~\ref{sec:ROM} describes the dimension reduction and the surrogate model for fire prediction. Section~\ref{sec:method} presents the methodology of VQ-VAE in generating spatial-temporal burned area snapshots. Numerical results, together with quantitative and qualitative analysis, are shown in Section~\ref{sec:results}. We finish the paper with 
a conclusion in Section~\ref{sec:conclusion}.

\section{Fire simulation and study area}
\label{sec:fire simulation}

\subsection{Study area}

We evaluate the performance of the generative and the surrogate models using a recent massive fire event in California, namely the Chimney fire in 2016. The exact ecoregion of this study is shown in Table~\ref{table:fire_areas}. We use active fire data from \ac{MODIS} and \ac{VIIRS} satellites. MODIS provides thermal observations  four times a day at a resolution of about 1km~(\cite{giglio2016collection}). VIIRS thermal data provides improved fire detection capabilities every 12 hours ~\cite{schroeder2014new}. In this study, the level 2 VIIRS I-band active fire product (VNP14IMG) with a resolution of 375 m is combined with the MODIS fire products at 1km to derive continuous daily fire perimeters, using the natural neighbour geospatial interpolation method~\cite{Scaduto2020}. 

\begin{table*}[h!]
\centering
\caption{Study areas of the three large wildfire events in California}
\begin{tabular}{ccccccccc} \toprule
    \textbf{Fire (Year)} & \multicolumn{2}{c}{\textbf{latitude}} & \multicolumn{2}{c}{\textbf{longitude}} &  \textbf{area}  \\ 
        \cmidrule{2-3}\cmidrule{4-5}
    & {{North}}
    & {{South}}
    & {{West}}
    & {{East}}
    \\ \midrule
    {Chimney (2016)}  & 37.7366 & 37.5093 & -119.9441 &  -119.7053 &  {$\approx246\textrm{km}^2$}  \\
    \bottomrule
\end{tabular}
\label{table:fire_areas}
\end{table*}

\subsection{Cellular Automata fire simulation}

\begin{figure}[h!]
\centering
\includegraphics[width=0.45\textwidth]{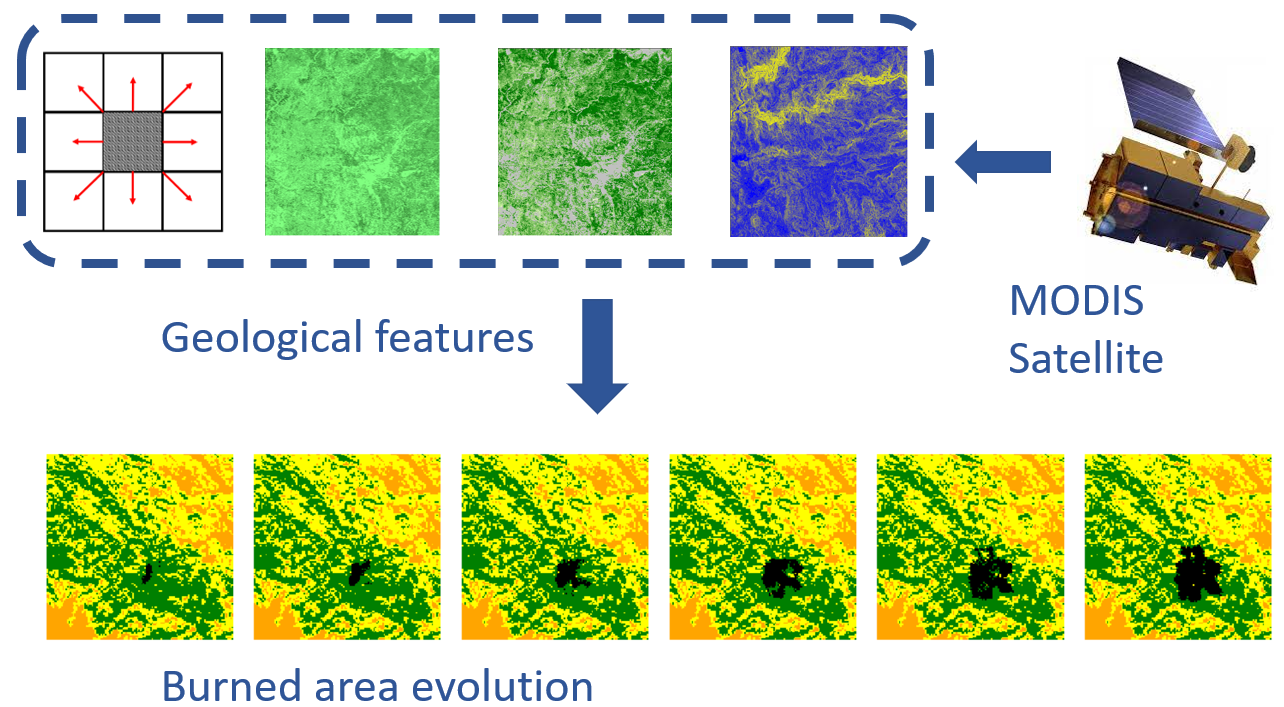}
\caption{Cellular Automata wildfire simulations}
\label{fig:CA}
\end{figure}

For the training dataset of the generative model and the surrogate model, we make use of an operational CA model~\cite{ALEXANDRIDIS2008}, which has been tested in the Spetses fire in 1990 in Greece~\cite{ALEXANDRIDIS2008} and several recent wildfire events in California~\cite{Cheng2022JCP,cheng2022parameter}.
As shown in Figure~\ref{fig:CA}, the model uses square meshes to simulate the random spatial spread of wildfires. The use of regular square meshes reduces the computational cost for large fire events compared to unstructured meshes. Four states are assigned to represent a cell at a discrete time, i.e., unburnable, unburned, burning and burned. When a neighbouring cell is burning, the transition from unburned to burning is random following a probabiliy determined by local geological variables. 
The fire propagation towards 8 neighbouring cells (as illustrated in Figure~\ref{fig:CA}) at a discrete time follows the probability,
\begin{align}
     P_\textrm{bun} = p_h (1+p_\textrm{veg}) (1+p_\textrm{den}) p_s p_w
     \label{eq:CA}
\end{align}
where  $p_\textrm{veg}$, $p_\textrm{den}$ , $p_s$ and $p_w$ are related to the local canopy density, canopy cover, landscape slope and wind speed/direction of the receiving cell, respectively~\cite{ALEXANDRIDIS2008}. 
These geological fields for the corresponding study areas are processed using remote sensing images of the MODIS satellite. These data are available at the Interagency Fuel Treatment Decision Support System (IFTDSS)~\cite{drury2016interagency}. The simulated burned area states are denoted as $\bx_t$ where $t$ is the time index. The outputs of the \ac{CA} simulations are used as training data for the generative and the surrogate models in this study.

\section{Reduced order surrogate modelling for burned area}
\label{sec:ROM}
In this section, we introduce the reduced order surrogate model that is coupled with the generative model in this paper. 

\subsection{Proper orthogonal decomposition for dimension reduction}

In this study, a set of $n_\textrm{state}$ state snapshots, from several \ac{CA} simulations, are represented by a tensor $\bX \in \mathbb{R}^{\textrm{dim}(\bx) \times n_\textrm{state}}$ where each column of $\bX$ represents a flattened burned area state at a given time step, that is,
\begin{align}
    \bX = \big[ \bx_0 \big| \bx_1 \big| ...\big| \bx_{n_\textrm{state}-1} \big].
\end{align}

The empirical covariance $\bC_{\bx}$ of $\bX$ can be written and decomposed as 

\begin{align}
    \bC_{\bx} = \frac{1}{n_\textrm{state}-1} \bX \bX^T = {\bL}_{\bX} {\bD}_{\bX} {{\bL}_{\bX}}^T
\end{align}
where the columns of ${\bL}_{\bX}$ are the principal components of $\bX$ and ${\bD}_{\bX}$ is a diagonal matrix with the associated eigenvalues $\{ \lambda_{\bX,i}, i=0,...,n_\textrm{state}-1\}$ in a decreasing order,
\begin{align}
  {\bD}_{\bX} =
  \begin{bmatrix}
    \lambda_{\bX,0} & & \\
    & \ddots & \\
    & & \lambda_{\bX,n_\textrm{state}-1}
  \end{bmatrix}.
\end{align}

For a truncation parameter $q \leq n_\textrm{state}$, a projection operator ${\bL}_{\bX,q}$ can be constructed by keeping the first $q$ columns of ${\bL}_{\bX}$. This projection operator can be obtained by a \ac{SVD}~\cite{baker2005singular} which does not require the computation of the full covariance matrix $\bC_{\bx}$. 
For a flattened field of burned area $\bx_t$, the compressed latent vector $\tilde{\bx}_t$ can be written as
\begin{align}
    \tilde{\bx}_t =  \textrm{Tr}({{\bL}_{\bX,q}}) \bx_t, \label{eq: reconstruction}
\end{align}
which is a reduced rank approximation to the full state vector $\bx_t$. $\textrm{Tr}(.)$ in Equation~\eqref{eq: reconstruction} denotes the transpose operator. 

The latent variable $\tilde{\bx}_t$ can be decompressed to a full vector $\bx_t^r$ by 
\begin{align}
    \bx_t^r = {{\bL}_{\bX,q}} \tilde{\bx}_t = {{\bL}_{\bX,q}} \textrm{Tr}({{\bL}_{\bX,q}}) \bx_t.
\end{align}
The compression rate $\rho_{\bx}$ and the compression accuracy $\gamma_{\bx}$ can be defined as:
\begin{align}
    \gamma_{\bx} = \sum_{i=0}^{q-1} \lambda^2_{\bX,i} \Big/ \sum_{i=0}^{n_\textrm{state}-1} \lambda^2_{\bX,i}  \quad \textrm{and} \quad \rho_{\bx} = q \big/ n_\textrm{state}. \label{eq:POD rate}
\end{align}
By reducing the space dimension, the surrogate model becomes more efficient and faster to train, but it may also lead to a loss of information.
Therefore, it is important to strike a balance between the dimensionality reduction and the preservation of the useful information.

\subsection{Predictive model using recurrent neural networks}

Once the data compression is performed, we aim to emulate the dynamics of fire spread in the low-dimensional latent space. 
In this study, we aim to perform sequence-to-sequence predictions with $m_{\textrm{in}}$ time steps as input and $m_{\textrm{out}}$ time steps as output. As pointed out by~\cite{Cheng2022JSC}, sequence-to-sequence predictions can decrease the online computational time, and more importantly, reduce the accumulation of prediction error. The time steps in \ac{LSTM} are determined by the \ac{CA} model introduced in Section~\ref{sec:fire simulation}.
Given a sequence of encoded variables $\tilde{\bX}_{\textrm{train}} = [\tilde{\bx}^{\textrm{train}}_1,\tilde{\bx}^{\textrm{train}}_2,..., \tilde{\bx}^{\textrm{train}}_{T_{\textrm{train}}}]$, the training of \ac{LSTM} can be performed by shifting the initial time step, i.e., 
\begin{align}
&[\tilde{\bx}^{\textrm{train}}_1,\tilde{\bx}^{\textrm{train}}_2,..., \tilde{\bx}^{\textrm{train}}_{m_{in}}] \overset{\textrm{LSTM} 
\textrm{train}}{\parbox{2.5cm}{\rightarrowfill}}\\& [\tilde{\bx}^{\textrm{train}}_{m_{in}+1},\tilde{\bx}^{\textrm{train}}_{m_{in}+2},..., \tilde{\bx}^{\textrm{train}}_{m_{in}+m_{out}}], \notag \\
&[\tilde{\bx}^{\textrm{train}}_2,\tilde{\bx}^{\textrm{train}}_3,..., \tilde{\bx}^{\textrm{train}}_{m_{in}+1}] \overset{\textrm{LSTM}  \textrm{train}}{\parbox{2.5cm}{\rightarrowfill}} \\& [\tilde{\bx}^{\textrm{train}}_{m_{in}+2},\tilde{\bx}^{\textrm{train}}_{m_{in}+3},..., \tilde{\bx}^{\textrm{train}}_{m_{in}+m_{out}+1}], \notag \\
& \vdots \notag \\
&[\tilde{\bx}^{\textrm{train}}_{T_\textrm{train}-m_{in}-m_{out}+1},..., \tilde{\bx}^{\textrm{train}}_{T_\textrm{train}-m_{out}}] \overset{\textrm{LSTM}  \textrm{train}}{\parbox{2.5cm}{\rightarrowfill}} \\& [\tilde{\bx}^{\textrm{train}}_{T_{\textrm{train}}-m_{out}+1},..., \tilde{\bx}^{\textrm{train}}_{T_{\textrm{train}}}]. \notag
\label{eq:train_LSTM}
\end{align}
Different metrics, such as \ac{MSE} or \ac{MAE}, can be used as training loss functions by measuring the mismatch between predicted and true latent variables.
As for the online prediction for a given test sequence $\tilde{\bX}_{\textrm{test}} = [\tilde{\bx}^{\textrm{test}}_1,\tilde{\bx}^{\textrm{test}}_2,..., \tilde{\bx}^{\textrm{test}}_{m_{in}}]$, an iterative process can be carried out for long-term forecasting, that is,
\begin{align}
&[\tilde{\bx}^{\textrm{test}}_1,\tilde{\bx}^{\textrm{test}}_2,..., \tilde{\bx}^{\textrm{test}}_{m_{in}}] \overset{\textrm{LSTM}  \textrm{predict}}{\parbox{2.5cm}{\rightarrowfill}}\\& [\tilde{\bx}^{\textrm{test}}_{m_{in}+1},\tilde{\bx}^{\textrm{test}}_{m_{in}+2},..., \tilde{\bx}^{\textrm{test}}_{m_{in}+m_{out}}], \notag \\
&[\tilde{\bx}^{\textrm{test}}_{m_{in}+1},\tilde{\bx}^{\textrm{test}}_{m_{in}+2},..., \tilde{\bx}^{\textrm{test}}_{m_{in}+m_{out}}] \overset{\textrm{LSTM} \textrm{predict}}{\parbox{2.5cm}{\rightarrowfill}}  \\ &[\tilde{\bx}^{\textrm{test}}_{m_{in}+m_{out}+1},..., \tilde{\bx}^{\textrm{test}}_{m_{in}+2m_{out}}]. \notag \\
& \vdots  \notag
\end{align}

\begin{figure}[h!]
\centering
\includegraphics[width=0.45\textwidth]{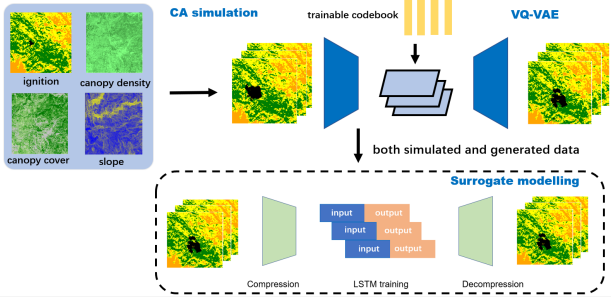}
\caption{Flowchart of the proposed generative approach for wildfire simulations}
\label{fig:flowchart_generative}
\end{figure}

\section{VQ-VAE for fire data generation}
\label{sec:method}
In this section, we describe how 3D VQ-VAE can be used to generate spatial-temporal consistent fire spread dynamics. The pipeline is illustrated in Figure~\ref{fig:flowchart_generative}.

\acp{VAE} are feed-forward neural networks belonging to unsupervised learning, being widely used for data compression and generation~\cite{kingma2013auto}. 
A typical \ac{VAE} consists of an encoder $\mathcal{E}$ which maps the input $\bx_t$ to a Gaussian distribution in the latent space $\mathcal{N}(\mu_t,\sigma_t^2)$, where $\mu_t, \sigma_t$ denote the mean and the standard deviation, respectively. 
A latent vector $\mathbf{z}_t$ can then be resampled and a generated vector $\hat{\bx}_t$ in the full space can be obtained through a decoder $\mathcal{D}$, 
\begin{align}
    \mathbf{z}_t = \mu_t + \epsilon_t \cdot \sigma_t \textrm{, where } \epsilon_t \sim \mathcal{N}(0,\mathbf{I}_{\textrm{dim}(\bz)}) \quad \textrm{and} \quad \hat{\mathbf{x}}_t = D(\mathbf{z}_t),
    \label{eq:reparameterization}
\end{align}
where $\mathbf{I}_{\textrm{dim}(\bz)}$ denotes the identity matrix in the latent space. The encoder $\mathcal{E}$ and the decoder $\mathcal{D}$ are then \rev{trained jointly through a \ac{MSE} reconstruction loss which measures the difference between $\bx_t$ and $\hat{\bx}_t$},
\begin{align}
    \rev{L_\text{rec} = ||\mathbf{x}_t-\hat{\mathbf{x}}_t||_F^2},
\end{align}
\rev{where $||\cdot ||_F$ denotes the Frobenius norm.}

\rev{A \ac{KLD} loss $L_\text{KLD}$ is also constructed with the idea of  approximating conditional likelihood distribution $p_{\theta}(\mathbf{z}_t|{\mathbf{x}}_t)$ by the posterior distribution $q_{\phi}(\mathbf{z}_t|{\mathbf{x}}_t)$} \cite{kingma2013auto},
\rev{\begin{align}
    L_\text{KLD} & =\mathbb{E}_{\mathbf{z}_t \sim q_\phi(\cdot \mid \mathbf{x}_t)}\left[\ln \frac{q_\phi(\mathbf{z}_t \mid \mathbf{x}_t)}{p_\theta(\mathbf{z}_t \mid \mathbf{x}_t)}\right] \\
& =\ln p_\theta(\mathbf{x}_t)+\mathbb{E}_{\mathbf{z}_t \sim q_\phi(\cdot \mid \mathbf{x}_t)}\left[\ln \frac{q_\phi(\mathbf{z}_t \mid \mathbf{x}_t)}{p_\theta(\mathbf{x}_t, \mathbf{z}_t)}\right]
\end{align}}

\rev{where $\phi$ and $\theta$ are the parameterization of the encoder $\mathcal{E}$ and the decoder $\mathcal{D}$, respectively.}
The \ac{KLD} loss regularizes the approximate posterior distribution over the latent variables to be similar to the prior distribution~\cite{hershey2007approximating}, ensuring the \ac{VAE} to generate diverse and meaningful samples.

Despite its great success in image compression and generation~\cite{bank2020autoencoders}, it has been noticed that standard \ac{VAE} is
not typically well suited for generating spatial-temporal data~\cite{liu2020photo}.
In fact, a standard \ac{VAE} assumes that the data at each time step is generated independently, leading to poor performance and unrealistic generated samples when being applied to spatial-temporal systems~\cite{fraccaro2016sequential}. To overcome these drawbacks, we use the VQ-VAE~\cite{van2017neural} which compresses the data points into a discretized
latent space instead of a continuous one.
 VQ-VAE can better capture the underlying structure of the data, leading to more coherent and structured generated samples compared to a standard \ac{VAE}~\cite{yan2021videogpt}. Furthermore, \rev{the discretization in the latent space in VQ-VAE encourages the model to focus on the main features such as objects and background, and implicitly alleviates the overfitting issue from the VAE-based model}.
 In this work, following the idea in~\cite{yan2021videogpt}, we apply the sequential VQ-VAE to encode a sequence of burned area snapshots $\bX_t^{t+m} = [\bx_t, \bx_{t+1},...,\bx_{t+m}]$
 into a series of latent vectors $\bZ_t^{t+m} = [\bz_t, \bz_{t+1},...,\bz_{t+m}]$, i.e.,
 \begin{align}
     \bZ_t^{t+m} = \mathcal{E} (\bX_t^{t+m}).
 \end{align}

The standard VQ-VAE uses $2D$ Convolutional layer to process images. To enable the VQ-VAE to process the sequential wildfire snapshots, we adopt $3D$ Convolutional layers in $\mathcal{E}$ and $\mathcal{D}$ by building a temporal axis $\mathcal{T}$. The wildfire snapshots are concatenated along $\mathcal{T}$ to form $3D$ inputs. VQ-VAE then discretize the latent code $\bZ_t^{t+m}$ in a codebook (an ensemble of latent sequences) $\mathcal{C} = \{\be_i\}_{i=1}^{\textrm{dim}(\bz)}$ where each element $\be_i$ consists of a sequence of latent vectors of the same dimension as $\bZ_t^{t+m}$. We then obtain the latent code $\be$ of $\bX_t^{t+m}$ through the nearest neighbour algorithm, that is,
\begin{align}
    \be = \underset{{\be_i \in \mathcal{C}}}{\argmin} \Big( ||\bZ_t^{t+m}-\be_i||_F \Big),
\end{align}
 The generated burned area sequences $\hat{\bX}_t^{t+m}$ can then be obtained through the decoder,
\begin{align}
    \hat{\bX}_t^{t+m} = \mathcal{D} (e).
\end{align}

It is worth mentioning that the codebook $\mathcal{C}$ in VQ-VAE is also trainable. It is trained jointly with $\mathcal{E}$ and $\mathcal{D}$. Following the idea of~\cite{yan2021videogpt}, three loss functions $L_{\textrm{recon}}, L_{\textrm{codebook}},L_{\textrm{commit}}$ are used in this study.  Similar to \ac{VAE}, the reconstruction loss $L_{\textrm{recon}}$ measures the mismatch between ${\bX}_t^{t+m}$ and $\hat{\bX}_t^{t+m}$. The codebook loss $L_{\textrm{codebook}}$ measures the distance between $\bZ_t^{t+m}$ and the closest element in the codebook. A commit loss $L_{\textrm{commit}}$ is also added to stabilize the encoder output $\bZ_t^{t+m}$. 
These loss functions are defined respectively as
\begin{align}
    \mathcal{L}_{recon}& = ||\hat{\bX}_t^{t+m}-{\bX}_t^{t+m}||^{2}_{F}\\
    \mathcal{L}_{codebook}& = ||stg(\bZ_t^{t+m})-\be||^{2}_{F}\\
    \mathcal{L}_{commit}&= \beta||stg(\be)-\bZ_t^{t+m}||^{2}_{F}, \label{eq:VQloss}
\end{align}\\
where $stg(\cdot )$ denotes the stop-gradient function~\cite{chen2021exploring}.
It is used in the VQ-VAE loss function to prevent the gradients from flowing through the codebook elements during the backpropagation. $\beta$ is an empirical coefficient that regularizes the weight of $\mathcal{L}_{commit}$. 
Finally, the total loss of VQ-VAE reads:
\begin{equation}
    \mathcal{L}_{total} = \mathcal{L}_{reconstruct}+\mathcal{L}_{codebook}+\mathcal{L}_{commit}.
\end{equation}
Once the VQ-VAE is trained, Gaussian noises (see Equation~\eqref{eq:reparameterization}) are added in the latent to generate a new fire spread scenario $\hat{\bX}_t^{t+m}$. More precisely, 
\begin{align}
    \hat{\bX}_t^{t+m} = \mathcal{D} (\alpha e + (1-\alpha) \epsilon), \label{eq:VQgeneration}
\end{align}
where $\alpha \in [0,1]$ is a hyperparameter. It is worth mentioning that Equation~\eqref{eq:VQgeneration} is only used for data generation. It is not involved in the training process.
\begin{figure}[h!]
\centering
\includegraphics[width=0.5\textwidth]{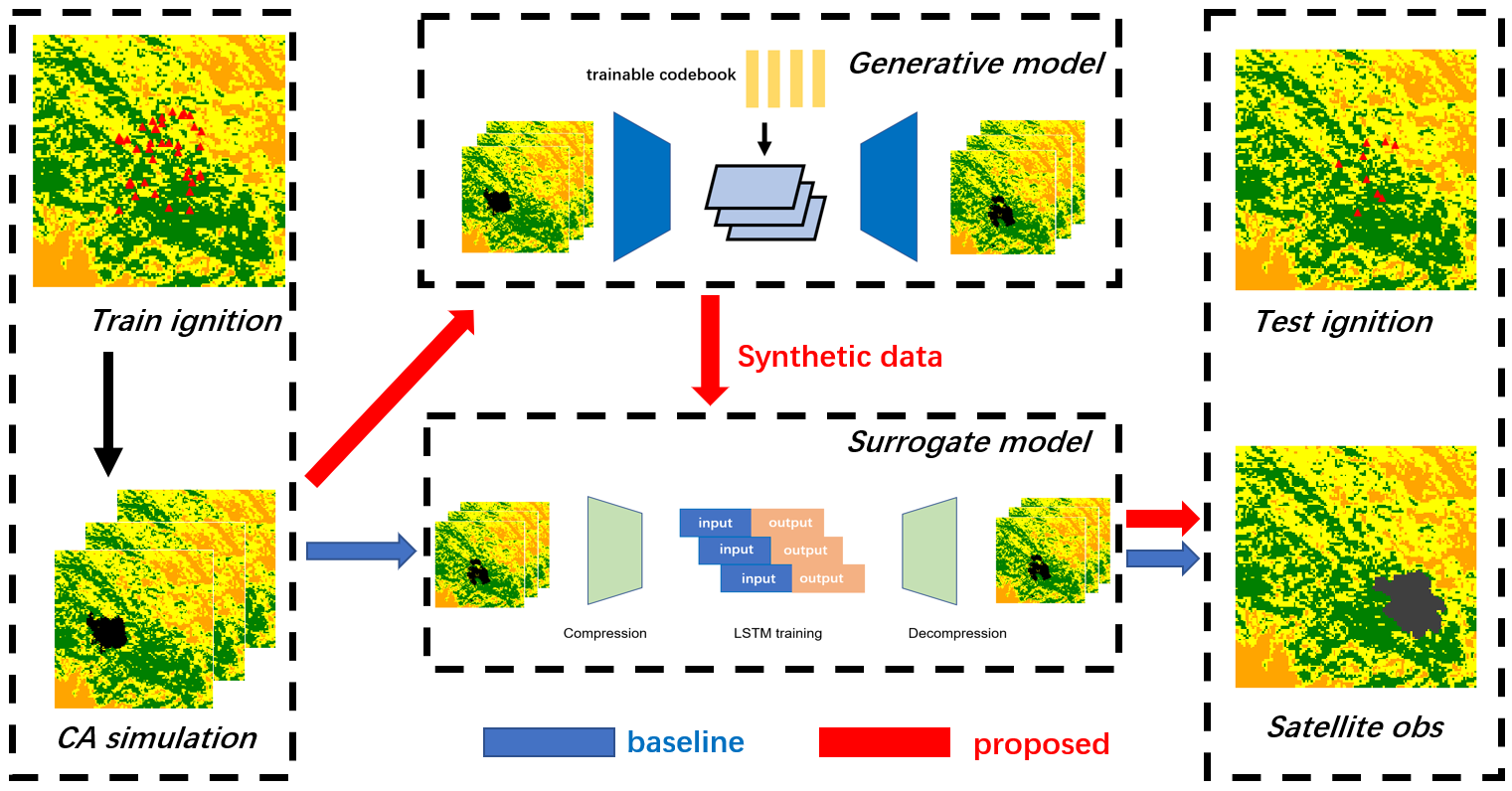}
\caption{Flowchart of the surrogate model training and testing using generated, simulated and satellite data. }
\label{fig:flowchart_Test}
\end{figure}

\section{Numerical Results}
\label{sec:results}
\begin{figure*}[ht!]
\centering
\makebox[\linewidth][c]{
\includegraphics[width = 1in]{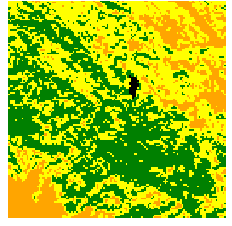}
\includegraphics[width = 1in]{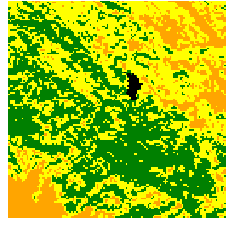}
\includegraphics[width = 1in]{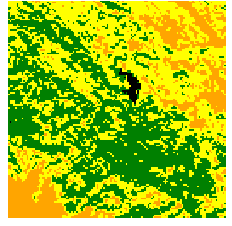}
\includegraphics[width = 1in]{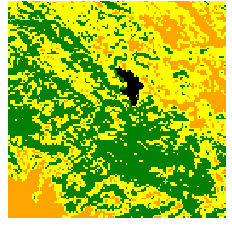}
\includegraphics[width = 1in]{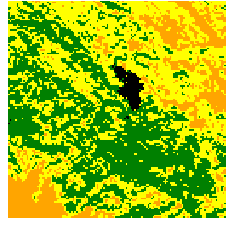}
\includegraphics[width = 1in]{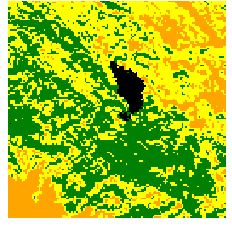}}\\
\makebox[\linewidth][c]{
\includegraphics[width = 1in]{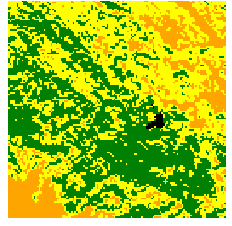}
\includegraphics[width = 1in]{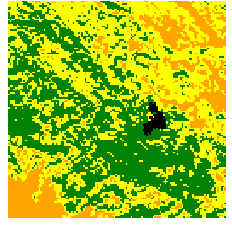}
\includegraphics[width = 1in]{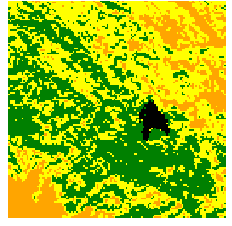}
\includegraphics[width = 1in]{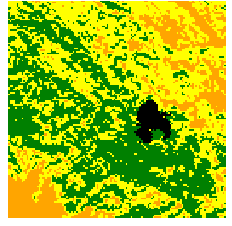}
\includegraphics[width = 1in]{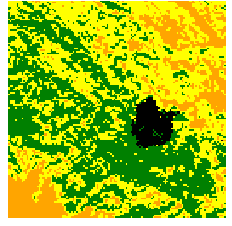}
\includegraphics[width = 1in]{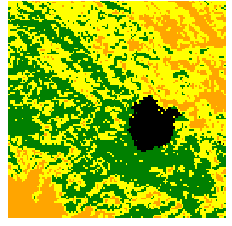}}\\
\makebox[\linewidth][c]{
\includegraphics[width = 1in]{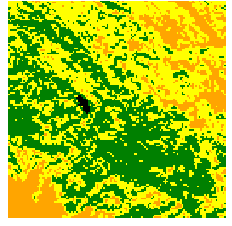}
\includegraphics[width = 1in]{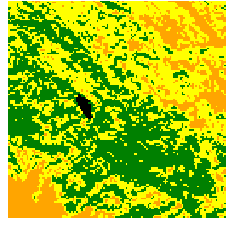}
\includegraphics[width = 1in]{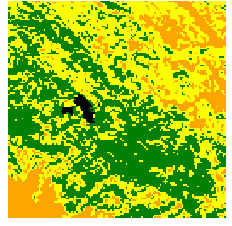}
\includegraphics[width = 1in]{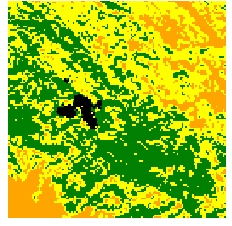}
\includegraphics[width = 1in]{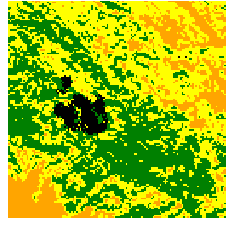}
\includegraphics[width = 1in]{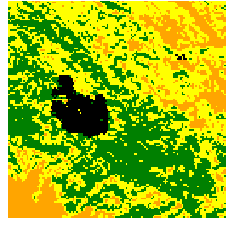}}\\
\makebox[\linewidth][c]{
\includegraphics[width = 1in]{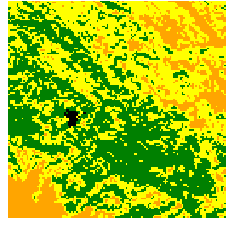}
\includegraphics[width = 1in]{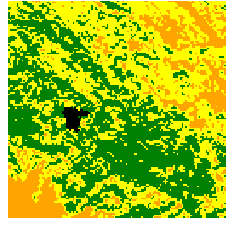}
\includegraphics[width = 1in]{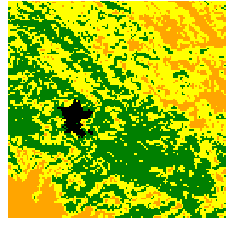}
\includegraphics[width = 1in]{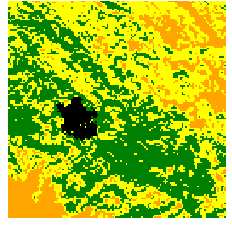}
\includegraphics[width = 1in]{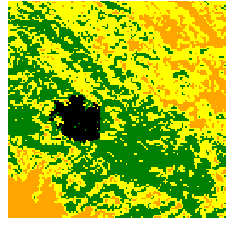}
\includegraphics[width = 1in]{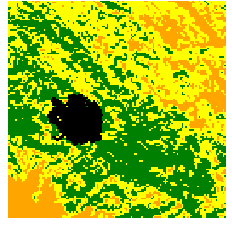}}\\
\makebox[\linewidth][c]{
\includegraphics[width = 1in]{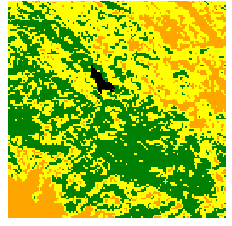}
\includegraphics[width = 1in]{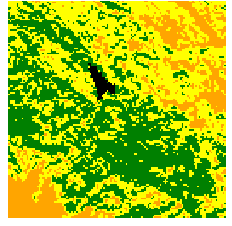}
\includegraphics[width = 1in]{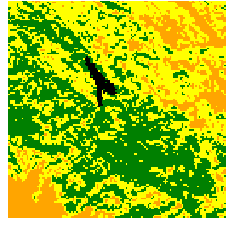}
\includegraphics[width = 1in]{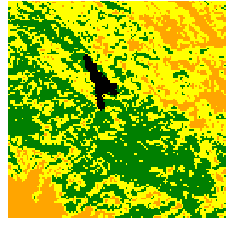}
\includegraphics[width = 1in]{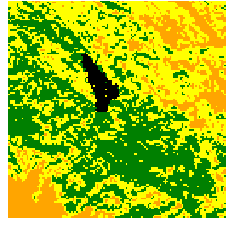}
\includegraphics[width = 1in]{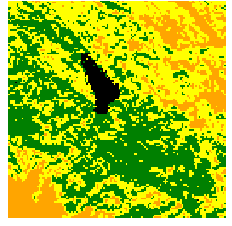}}\\
\makebox[\linewidth][c]{
\includegraphics[width = 1in]{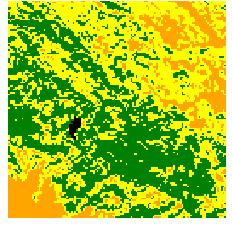}
\includegraphics[width = 1in]{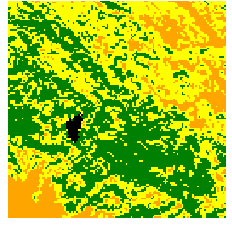}
\includegraphics[width = 1in]{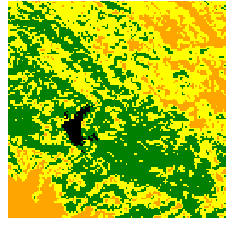}
\includegraphics[width = 1in]{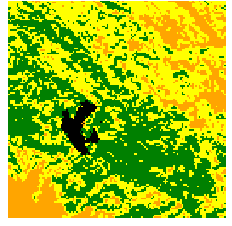}
\includegraphics[width = 1in]{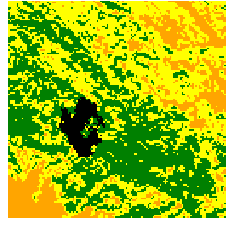}
\includegraphics[width = 1in]{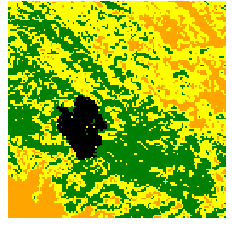}}\\
\makebox[\linewidth][c]{
\includegraphics[width = 1in]{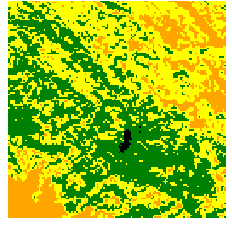}
\includegraphics[width = 1in]{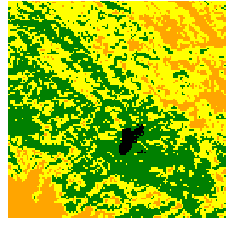}
\includegraphics[width = 1in]{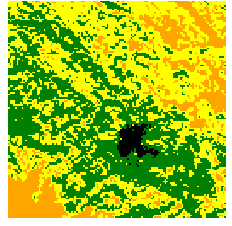}
\includegraphics[width = 1in]{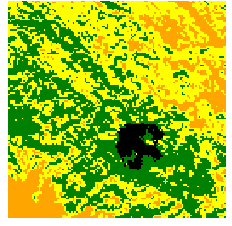}
\includegraphics[width = 1in]{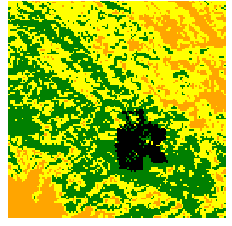}
\includegraphics[width = 1in]{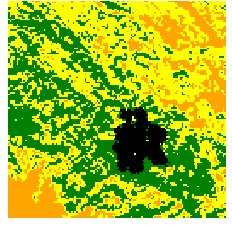}}\\
\makebox[\linewidth][c]{
\includegraphics[width = 1in]{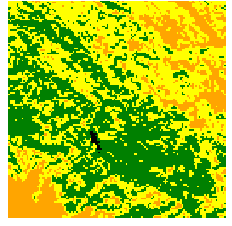}
\includegraphics[width = 1in]{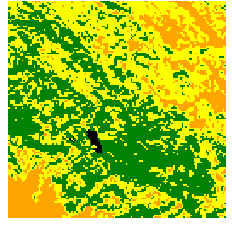}
\includegraphics[width = 1in]{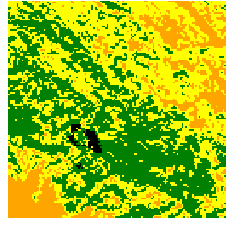}
\includegraphics[width = 1in]{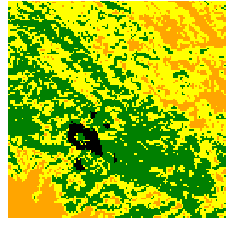}
\includegraphics[width = 1in]{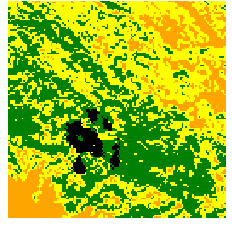}
\includegraphics[width = 1in]{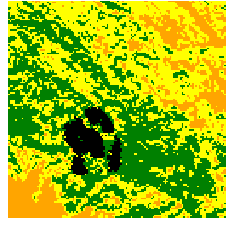}}\\
\caption{Generated sequences of burned area in the ecoregion of the \textit{Chimney} fire in California. Each row represents a generated fire event with snapshots at 12, 24, 36, 48, 60, 72 hours after fire ignition.}
\label{fig:generate}
\end{figure*}
\subsection{Experiments set up}

In this work, 40 \ac{CA} simulations are used as the training dataset for the VQ-VAE generative model. The ignition points are randomly chosen in the central zone of the ecoregion to avoid the boundary problem as shown in Figure~\ref{fig:flowchart_Test}. 
In this study, the parameters in VQ-VAE are fixed as $\beta = 0.25$ (in Equation~\eqref{eq:VQloss}) and $\alpha = 0.6$ (in Equation~\eqref{eq:VQgeneration}).
The performance of the generative model is evaluated through two aspects, namely the realsticness of the generated fire events regarding geological variables and the improvement of the predictive surrogate model which use the generated fire events as training data. For the latter, 8 \ac{CA} simulations, different from the training dataset are generated to find out if the surrogate models can correctly emulate the fire spread with unseen ignition points. For a fair comparison, two pipelines are established,
\begin{itemize}
    \item \textit{baseline}: The surrogate model (including both \ac{POD} and \ac{LSTM}) are trained only using the 40 \ac{CA} simulations in the training dataset.
    \item \textit{proposed}: 500 wildfire events generated through VQ-VAE and the 40 \ac{CA} simulations are used to train the surrogate model.
\end{itemize}
The training and testing processes of the two surrogate models are illustrated in Figure~\ref{fig:flowchart_Test}. The performance of the two surrogate models is then compared using the test dataset of the \ac{CA} simulations and satellite data of the Chimney wildfire event.
In the training and the test datasets, each \ac{CA} simulation consists of 16 snapshots where the time interval between two snapshots is equivalent to 6 hours in real time. 
Since the state is represented by integers in the \ac{CA} model (see Section~\ref{sec:fire simulation}), post-processing is required for generated/predicted values. Here, the threshold is chosen to be 0.4, i.e., the cells with a value higher than 0.4 are considered 'burned'. This threshold is set empirically following the previous study~\cite{Cheng2022JCP}.
For simplicity, the burned area state and the geological fields (vegetation and slope) are all resized to $128\times128$ pixels. 
The generative model and the surrogate model are trained on a work station with an NVIDIA RTX A6000 GPU and 48 GB Memory. As for the online generation/simulation, VQ-VAE, \ac{CA} and \ac{MTT} are all performed on a same labtop CPU of Intel(R) Core(TM) i7-10810U with 16 GB Memory.

\subsection{Generated wildfire events}

\begin{figure}[ht!]
\centering
\subfloat[Bured area vs. time]{\includegraphics[width=0.3\textwidth]{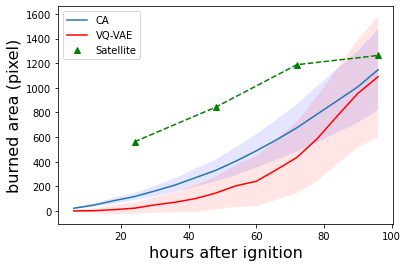}} \\
\subfloat[Bured area vs. vegetation]{\includegraphics[width=0.24\textwidth]{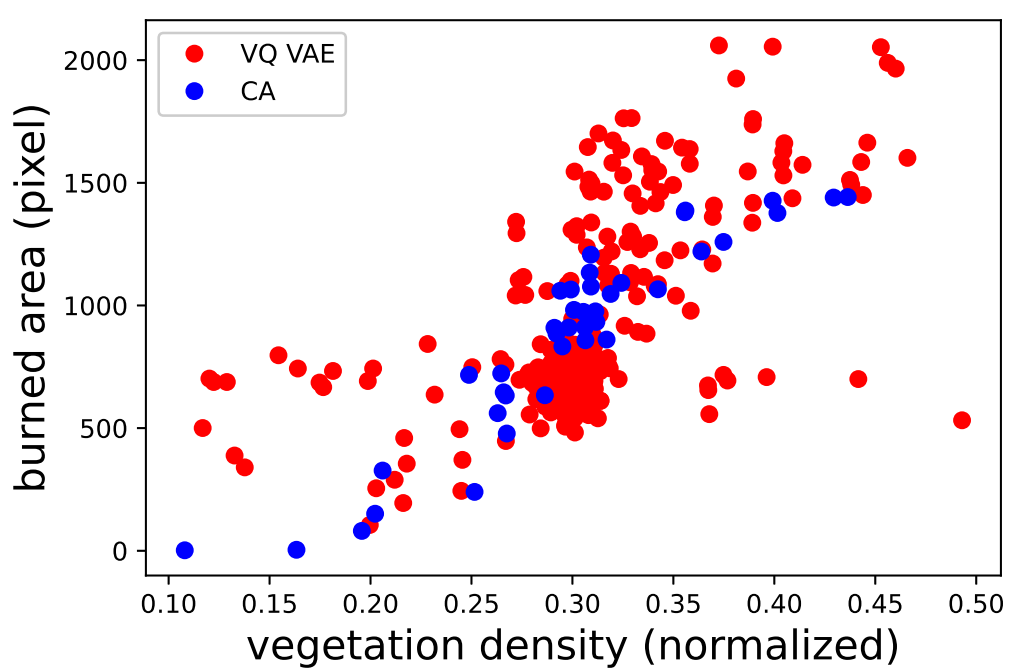}}
\subfloat[Bured area vs. slope]{\includegraphics[width=0.24\textwidth]{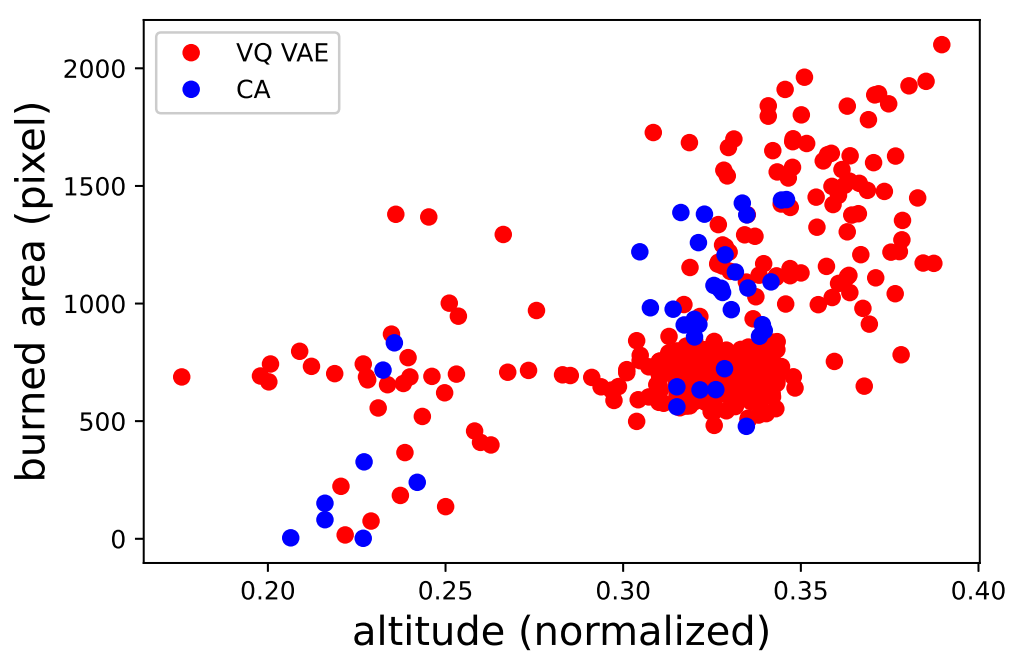}}
\caption{Comparison of generated and simulated wildfire events in terms of burned area in number of pixels regarding fire propagation time, vegetation density and slope.}
\label{fig:scatter}
\end{figure}

Figure~\ref{fig:generate} illustrates 8 generated fire events with random ignition points. For each generated fire, 6 snapshots respectively at 12, 24, 36, 48, 60, 72 hours after the fire ignition are shown. The background color map indicates the vegetation distribution where the green color refers to high vegetation density.  

The generated data present characteristics that are in alignment with observable phenomena in \ac{CA} simulations. First, the burned areas (i.e., number of burned pixels) are always increasing in each generated fire event. This observation is quantified in Figure~\ref{fig:scatter} (a) where we show the averaged burned pixels against the hours after fire ignition. Overall the generated (i.e., VQ-VAE) and the simulated data (i.e., \ac{CA}) demonstrate concurrent patterns despite that generated fire grow slightly slower in the nascent phase. The generated fire events also exhibit a larger standard deviation against propagation time, which is to be expected in this study. The exact burned areas observed from the satellite for the Chimney fire event are slightly larger than the simulated and generated burned areas. 

It can also been clearly seen from Figure~\ref{fig:generate} that the generated fires tend to burn out the cells with high vegetation density which appears to be consistent with the \ac{CA} propagation function (Equation~\eqref{eq:CA}). To verify this assertion, we plot in Figure~\ref{fig:scatter} (b,c), the final burned area (72 hours after the fire ignition) in number of cells against the averaged vegetation density and the slope elevation around the ignition point. More precisely, the ignition point of generated fire events is obtained by calculating the barycentre of the burned area 12 hours after ignition. The vegetation density and the slope elevation are normalized to the range of $[0,1]$. There is conclusive evidence in Figure~\ref{fig:scatter} that the generated fires show similar patterns compared to the \ac{CA} simulations in the training dataset. Thus, the impact of vegetation and slope on the evolution of the burned area has been well understood by the generative model.

It is worth mentioning that all training data are generated with a single ignition point for each fire scenario. However, as shown in Figure~\ref{fig:generate}, separate burned regions can be found in the synthetic data (especially the last fire sequence). More importantly, the growth of each burned region appears to be consistent with the reality.
In fact, multiple ignition points are frequently observed in actual wildfire occurrences. This phenomena demonstrates the capability of VQ-VAE in generalizing and extending \ac{CA} simulations with realistic scenarios.
The averaged computational time of generating/simulating fire spread of 8 days in the ecoregion of the Chimney fire is shown in Table~\ref{table:time}. VQ-VAE result in a 5 order of magnitude acceleration, compared to the \ac{CA} and \ac{MTT} fire simulators~\cite{finney2002fire}. 
\begin{table}
\centering
\caption{Averaged computational time for generating a fire spread of 8 days}
\begin{tabular}{ccc} \toprule
     {{VQ-VAE}}
    & {{\ac{CA}}}
    & {{\ac{MTT}}}
    \\ \midrule
    0.26s  & $\approx 35$ min &  $7$ min $11s$ \\
    \bottomrule
\end{tabular}
\label{table:time}
\end{table}
In summary, VQ-VAE allows to easily enlarge the simulation datasets with high-fidelity synthetic data. It shows great potential in generating realistic fire scenarios, which can be used to evaluate the area at risk.
The efficiency of VQ-VAE can also be crucial for data-demanding downstream tasks, such as surrogate model training. 

\subsection{Surrogate model}
We compare the surrogate model trained on both generated (500 sequences) and simulated (40 sequences) data (i.e., \textit{proposed} method in Figure~\ref{fig:flowchart_Test}) and the one trained only on simulation (40 sequences) data (i.e., \textit{baseline} method in Figure~\ref{fig:flowchart_Test}). The performance is evaluated on a test dataset consisting of 8 independent \ac{CA} simulations and the observed snapshots of the Chimney fire event. For each sequences of burned area, the first 4 snapshots, corresponding to 6, 12, 18, 24 hours after fire ignition, are used as input for the predictive surrogate model. The latter then predict the burned area for 30, 36, 42, 48 hours after fire ignition, as introduced in Section~\ref{sec:method}. For both surrogate models, the validation dataset consists of 8 independent \ac{CA} simulations, different from the training and the test data. 

\begin{figure}[ht!]
\centering
\subfloat[Validation loss]{\includegraphics[width=0.3\textwidth]{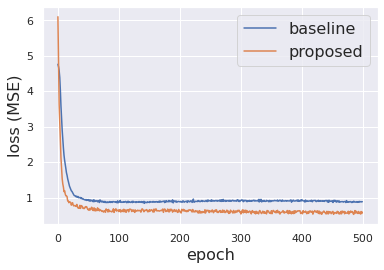}} \\
\subfloat[algorithm performance]{\includegraphics[width=0.38\textwidth]{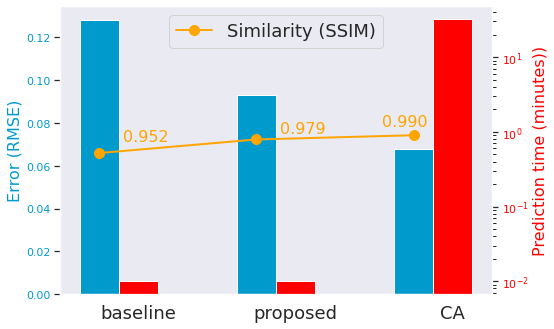}}
\caption{Comparison of \textit{baseline} and \textit{proposed} approaches in terms of validation loss (a) and prediction performance (b).}
\label{fig:loss}
\end{figure}

The evolution of the loss value (\ac{MSE}) against the number of training epochs is depicted in Figure~\ref{fig:loss} (a). It can be clearly observed that, despite some oscillations introduced by the generated data, the loss of the \textit{proposed} approach converges faster than \textit{baseline}, and reaches a lower stable value. Some quantitative metrics are used to assess the prediction performance as shown in Figure~\ref{fig:loss} (b). For comparison purpose, we also run the \ac{CA} from the burned area of 24 hours after ignition. As mentioned in~\ref{sec:fire simulation}, \ac{CA} is a probabilistic model, thus the re-ran simulation will lead to a different burned area as the one in the test dataset. The blue bar stands for the \ac{RMSE}, which represents the mismatch normalized by the exact burned area. The red bar illustrates the online prediction time and the orange line represents the \ac{SSIM}, which measures the similarity between the true burned area and the prediction. All metrics are obtained using the average score on the 8 fire sequences in the test dataset. Consistent with our analysis for Figure~\ref{fig:loss} (a), the \textit{proposed} approach shows a significant advantage in terms of both \ac{RMSE} and \ac{SSIM} compared to the \textit{baseline}. In fact, the accuracy of the \textit{proposed} surrogate model is close to a re-implementation of \ac{CA} with a speed-up of 4 orders of magnitude. 

\begin{figure}[h!]
  \centering
\subfloat[truth (30h)]{\includegraphics[width=0.125\textwidth]{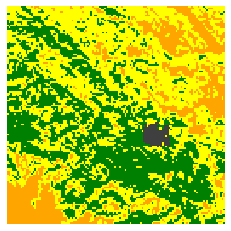}} 
\subfloat[truth (36h)]{\includegraphics[width=0.125\textwidth]{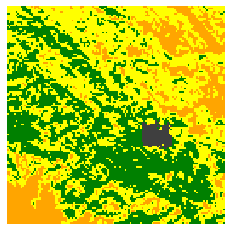}} 
\subfloat[truth (42h)]{\includegraphics[width=0.125\textwidth]{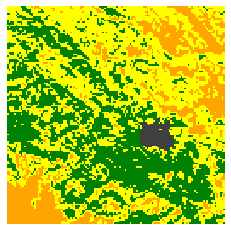}} 
\subfloat[truth (48h)]{\includegraphics[width=0.125\textwidth]{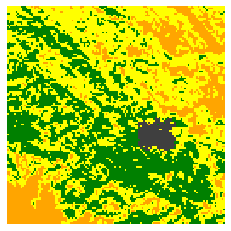}}\\
\subfloat[proposed (30h)]{\includegraphics[width=0.125\textwidth]{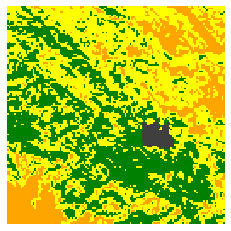}} 
\subfloat[proposed (36h)]{\includegraphics[width=0.125\textwidth]{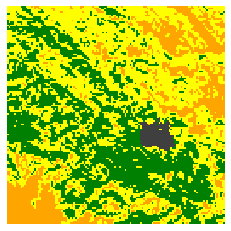}} 
\subfloat[proposed (42h)]{\includegraphics[width=0.125\textwidth]{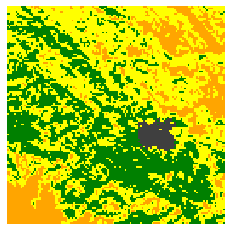}} 
\subfloat[proposed (48h)]{\includegraphics[width=0.125\textwidth]{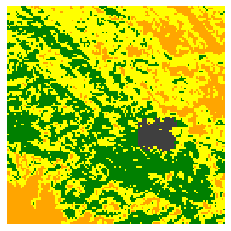}}\\
\subfloat[mismatch (30h)]{\includegraphics[width=0.125\textwidth]{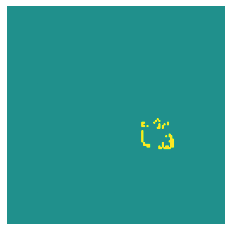}} 
\subfloat[mismatch (36h)]{\includegraphics[width=0.125\textwidth]{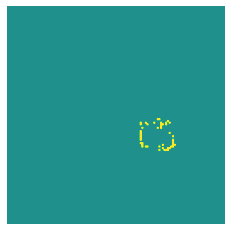}}
\subfloat[mismatch (42h)]{\includegraphics[width=0.125\textwidth]{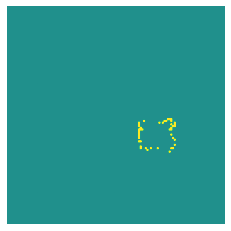}}
\subfloat[mismatch (48h)]{\includegraphics[width=0.125\textwidth]{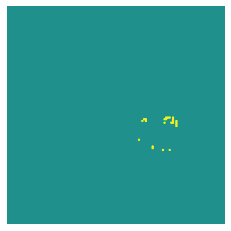}}\\
\subfloat[baseline (30h)]{\includegraphics[width=0.125\textwidth]{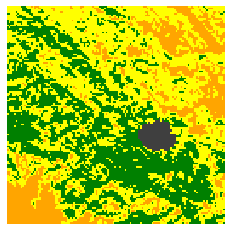}} 
\subfloat[baseline (36h)]{\includegraphics[width=0.125\textwidth]{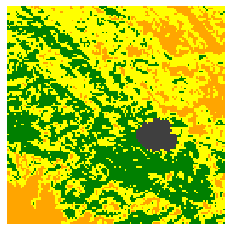}} 
\subfloat[baseline (42h)]{\includegraphics[width=0.125\textwidth]{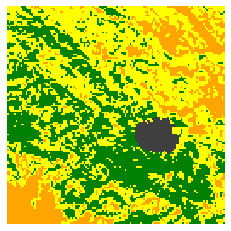}} 
\subfloat[baseline (48h)]{\includegraphics[width=0.125\textwidth]{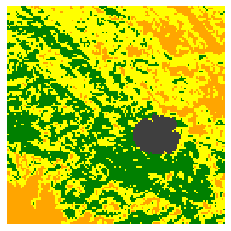}}\\
\subfloat[mismatch (30h)]{\includegraphics[width=0.125\textwidth]{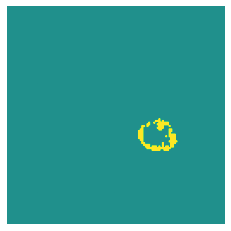}} 
\subfloat[mismatch (36h)]{\includegraphics[width=0.125\textwidth]{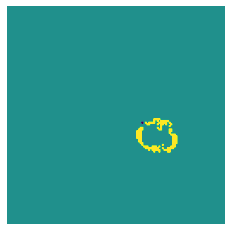}}
\subfloat[mismatch (42h)]{\includegraphics[width=0.125\textwidth]{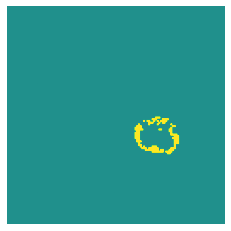}}
\subfloat[mismatch (48h)]{\includegraphics[width=0.125\textwidth]{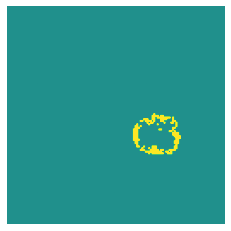}}\\
\caption{Fire spread prediction on test case 1 generated by \ac{CA}}
\label{fig:testCA1}
\end{figure}

\begin{figure}[h!]
  \centering
\subfloat[truth (30h)]{\includegraphics[width=0.125\textwidth]{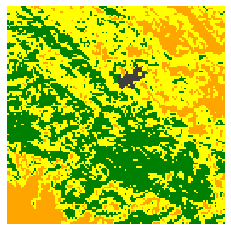}} 
\subfloat[truth (36h)]{\includegraphics[width=0.125\textwidth]{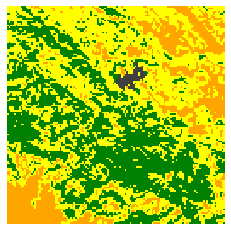}} 
\subfloat[truth (42h)]{\includegraphics[width=0.125\textwidth]{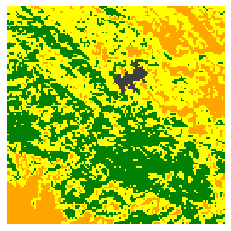}} 
\subfloat[truth (48h)]{\includegraphics[width=0.125\textwidth]{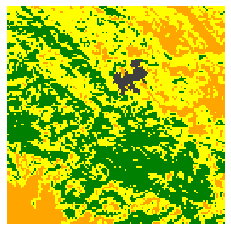}}\\
\subfloat[proposed (30h)]{\includegraphics[width=0.125\textwidth]{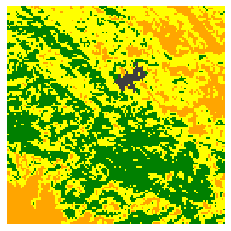}} 
\subfloat[proposed (36h)]{\includegraphics[width=0.125\textwidth]{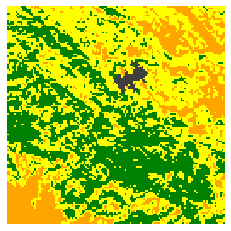}} 
\subfloat[proposed (42h)]{\includegraphics[width=0.125\textwidth]{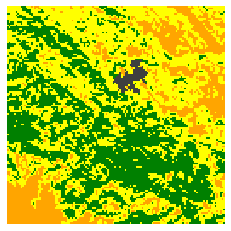}} 
\subfloat[proposed (48h)]{\includegraphics[width=0.125\textwidth]{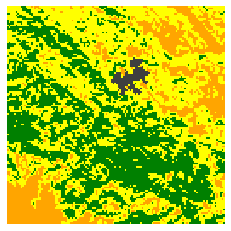}}\\
\subfloat[mismatch (30h)]{\includegraphics[width=0.125\textwidth]{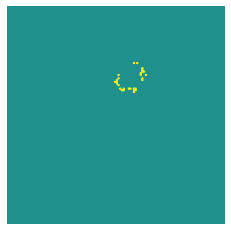}} 
\subfloat[mismatch (36h)]{\includegraphics[width=0.125\textwidth]{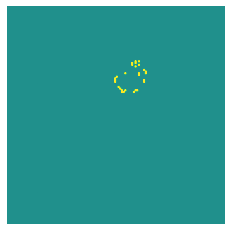}}
\subfloat[mismatch (42h)]{\includegraphics[width=0.125\textwidth]{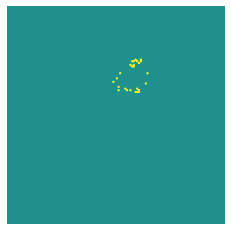}}
\subfloat[mismatch (48h)]{\includegraphics[width=0.125\textwidth]{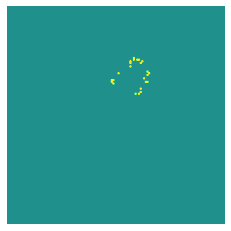}}\\
\subfloat[baseline (30h)]{\includegraphics[width=0.125\textwidth]{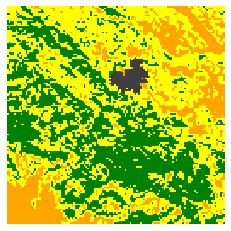}} 
\subfloat[baseline (36h)]{\includegraphics[width=0.125\textwidth]{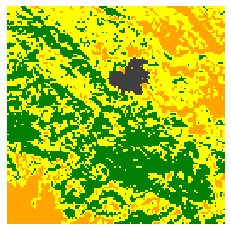}} 
\subfloat[baseline (42h)]{\includegraphics[width=0.125\textwidth]{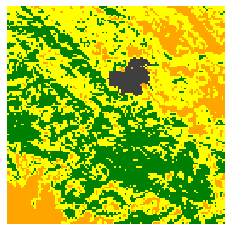}} 
\subfloat[baseline (48h)]{\includegraphics[width=0.125\textwidth]{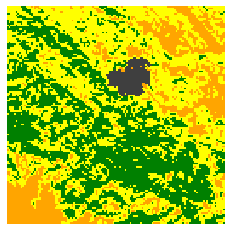}}\\
\subfloat[mismatch (30h)]{\includegraphics[width=0.125\textwidth]{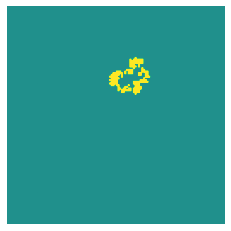}} 
\subfloat[mismatch (36h)]{\includegraphics[width=0.125\textwidth]{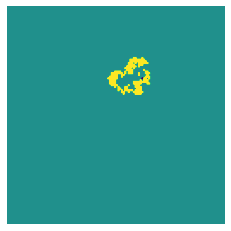}}
\subfloat[mismatch (42h)]{\includegraphics[width=0.125\textwidth]{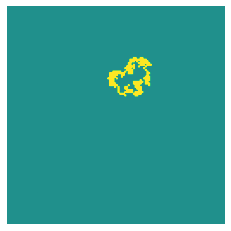}}
\subfloat[mismatch (48h)]{\includegraphics[width=0.125\textwidth]{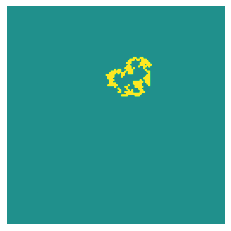}}\\
\caption{Fire spread prediction on test case 2 generated by \ac{CA}}
\label{fig:testCA2}
\end{figure}

Two fire events in the test data are illustrated in Figure~\ref{fig:testCA1} and~\ref{fig:testCA2}, respectively.  
In accordance with our findings in Figure~\ref{fig:loss}, the \textit{proposed} approach succeeds in delivering accurate predictions of the burned area. Only minor difference can be observed against the ground truth. The \textit{baseline} approach, on the other hand, manages to roughly predict the progression of fire dissemination but fails in specifying the precise geometry of burned area. This is mainly due to the lack of training data. Figure~\ref{fig:testCA2} shows a particular case where the vegetation density is small near the ignition point, leading to a deceleration of fire spread. This phenomenon is only well understood by the \textit{proposed} approach.

\begin{figure}[h!]
  \centering
\subfloat[obs (30h)]{\includegraphics[width=0.125\textwidth]{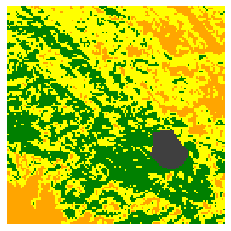}} 
\subfloat[obs (36h)]{\includegraphics[width=0.125\textwidth]{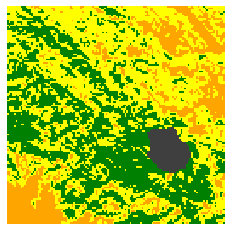}} 
\subfloat[obs (42h)]{\includegraphics[width=0.125\textwidth]{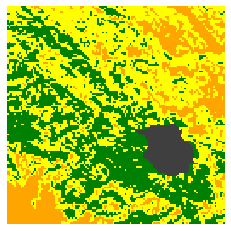}} 
\subfloat[obs (48h)]{\includegraphics[width=0.125\textwidth]{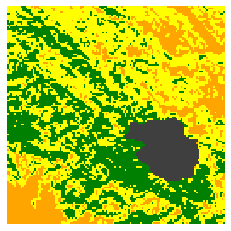}}\\
\subfloat[proposed (30h)]{\includegraphics[width=0.125\textwidth]{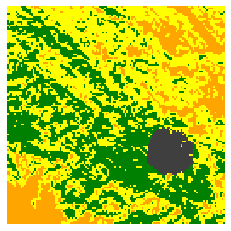}} 
\subfloat[proposed (36h)]{\includegraphics[width=0.125\textwidth]{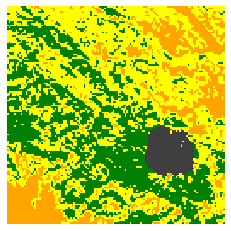}} 
\subfloat[proposed (42h)]{\includegraphics[width=0.125\textwidth]{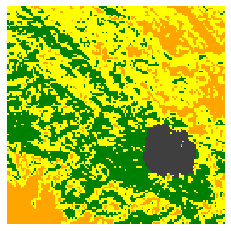}} 
\subfloat[proposed (48h)]{\includegraphics[width=0.125\textwidth]{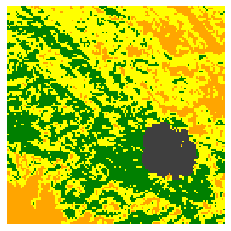}}\\
\subfloat[mismatch (30h)]{\includegraphics[width=0.125\textwidth]{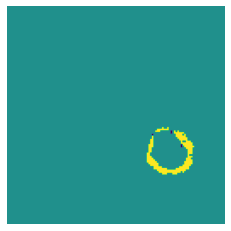}} 
\subfloat[mismatch (36h)]{\includegraphics[width=0.125\textwidth]{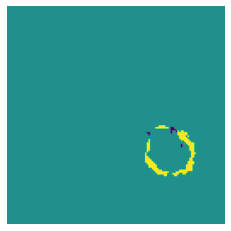}} 
\subfloat[mismatch (42h)]{\includegraphics[width=0.125\textwidth]{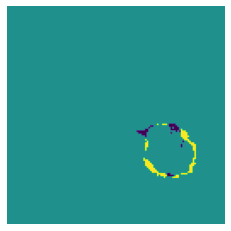}} 
\subfloat[mismatch (48h)]{\includegraphics[width=0.125\textwidth]{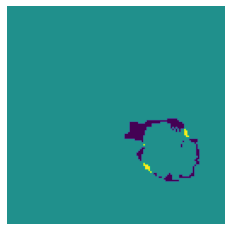}}\\
\subfloat[baseline (30h)]{\includegraphics[width=0.125\textwidth]{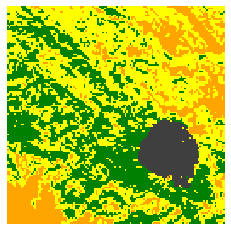}} 
\subfloat[baseline (36h)]{\includegraphics[width=0.125\textwidth]{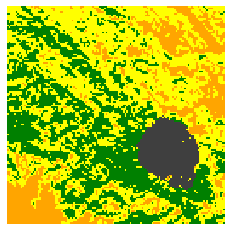}} 
\subfloat[baseline (42h)]{\includegraphics[width=0.125\textwidth]{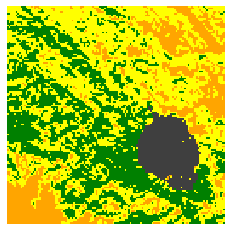}} 
\subfloat[baseline (48h)]{\includegraphics[width=0.125\textwidth]{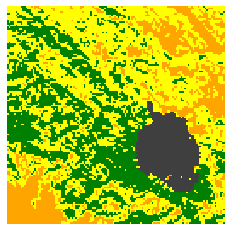}}\\
\subfloat[mismatch (30h)]{\includegraphics[width=0.125\textwidth]{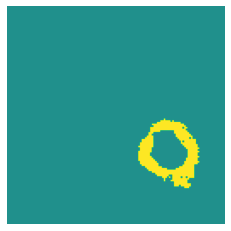}} 
\subfloat[mismatch (36h)]{\includegraphics[width=0.125\textwidth]{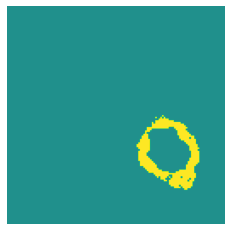}} 
\subfloat[mismatch (42h)]{\includegraphics[width=0.125\textwidth]{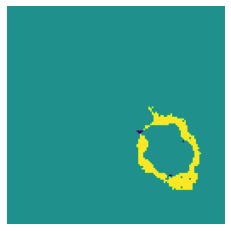}} 
\subfloat[mismatch (48h)]{\includegraphics[width=0.125\textwidth]{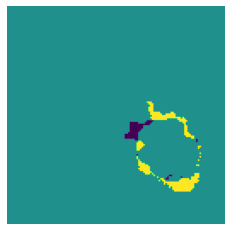}}\\
\caption{Fire spread prediction for the real Chimney wildfire event}
\label{fig:testobs}
\end{figure}

We also test the prediction performance of surrogate models applied on the Chimney fire event (Table~\ref{table:fire_areas}), which took place on August 13, 2016. The observed burned area is provided by the MODIS satellite as described in Section~\ref{sec:fire simulation}. Since the observation data are on daily basis~\cite{Cheng2022JCP}, image interpolation method proposed by~\cite{schenk2000efficient} is applied to emulate the burned area at intervals of 6 hours. As shown in Figure~\ref{fig:testobs}, for both surrogate models larger area of mismatch can be noticed compared to the \ac{CA} test data (Figure~\ref{fig:testCA1},~\ref{fig:testCA2}). However, a substantial gain of the \textit{proposed} approach can be found against the \textit{baseline}. In a consistent manner, the averaged \ac{RMSE} for \textit{proposed} and \textit{baseline} approaches are $28.4\%$ and $18.9\%$, respectively.

These numerical results convincingly demonstrate the strength of including generated data in the surrogate model to improve the robustness and generalizability when predicting unseen scenarios.

\section{Conclusion}
\label{sec:conclusion}

One of the main challenges in the current wildfire forecasting study is that physics-based simulation processes incur a significant computational cost and demand a substantial allocation of time.
In this study, we use 3D VQ-VAE for generating spatial-temporal consistent and physically realistic wildfire scenarios in a given ecoregion.  
This work is a first attempt to generate sequential fire spread instances via a generative deep learning model. 
Our approach is tested using data from a recent massive wildfire event in California. The proposed approach manages to generate wildfire spread of 8 days in 0.3 seconds, resulting in a speed-up of 4 orders of magnitude compared to the state-of-the-art \ac{CA} model. The generated data can be used to assess the likelihood of high fire susceptibility in the given ecoregion. Furthermore, numerical results show that the accuracy and the generalizability of a fire predictive model can be significantly enhanced by including generated fire scenarios in the training dataset.

As a proof of concept, initial fire simulations are produced from \ac{CA} as an example of physics-based models. A future investigation calls for additional work to incorporate more advanced fire behavior models such as FARSITE~\cite{finney1998farsite} and SPARK~\cite{hilton2020radiant}, which take into account, for instance, surface-to-crown fire transition~\cite{weise2018surface} and fire spotting. Future work can also be considered to combine the generative and predictive models in an end-to-end framework.

\section*{Acronyms}

\begin{acronym}[AAAAA]
\footnotesize{
\acro{NN}{Neural Network}
\acro{DNN}{Deep Neural Network}
\acro{ML}{Machine Learning}
\acro{MAE}{Mean Absolute Error}
\acro{LA}{Latent Assimilation}
\acro{DA}{Data Assimilation}
\acro{PR}{Polynomial Regression}
\acro{AE}{Autoencoder}
\acro{VAE}{Variational Autoencoder}
\acro{CAE}{Convolutional Autoencoder}
\acro{VAE}{Variational Autoencoder}
\acro{SVM}{Support Vector Machine}
\acro{MTT}{Minimum Travel Time}
\acro{BLUE}{Best Linear Unbiased Estimator}
\acro{3D-Var}{3D Variational}
\acro{GAN}{Generative Adversarial Network}
\acro{RNN}{Recurrent Neural Network}
\acro{CNN}{Convolutional Neural Network}
\acro{SSIM}{Structural Similarity Index Measure}
\acro{LSTM}{long short-term memory}
\acro{POD}{Proper Orthogonal Decomposition}
\acro{PCA}{Principal Component Analysis}
\acro{PC}{principal component}
\acro{SVD}{Singular Value Decomposition}
\acro{ROM}{reduced-order modelling}
\acro{CFD}{computational fluid dynamics}
\acro{1D}{one-dimensional}
\acro{2D}{two-dimensional}
\acro{NWP}{numerical weather prediction}
\acro{RMSE}{Relative Mean Square Error}
\acro{MSE}{Mean Square Error}
\acro{S2S}{sequence-to-sequence}
\acro{R-RMSE}{relative root mean square error}
\acro{BFGS}{Broyden–Fletcher–Goldfarb–Shanno}
\acro{LHS}{Latin Hypercube Sampling}
\acro{AI}{artificial intelligence}
\acro{DL}{Deep Learning}
\acro{PIV}{Particle Image Velocimetry}
\acro{LIF}{Laser Induced Fluorescence}
\acro{KNN}{K-nearest Neighbours}
\acro{DT}{Decision Tree}
\acro{RF}{Random Forest}
\acro{KF}{Kalman filter}
\acro{CART}{Classification And Regression Tree}
\acro{CA}{Cellular Automata}
\acro{MLP}{Multi Layer Percepton}
\acro{GLA}{Generalised Latent Assimilation}
\acro{3Dvar}{Three-dimensional  variational data assimilation }
\acro{4Dvar}{Four-dimensional  variational data assimilation }
\acro{KLD}{Kullback–Leibler Divergence }
\acro{MODIS}{Moderate Resolution Imaging Spectroradiometer}
\acro{VIIRS}{Visible Infrared Imaging Radiometer Suite}}
\end{acronym}

\section*{\rev{Appendix: parameter studies}}

\rev{Here we perform additional numerical tests for parameter studies, regarding the value of $\alpha$ and $\beta$ (Equation \eqref{eq:VQloss} and \eqref{eq:VQgeneration}), and the number of generated videos used in training the surrogate model}

\subsection{VQ-VAE parameters: $\alpha$ and $\beta$}

\begin{figure}[h!]
  \centering
\subfloat[$\alpha=0.3$ (30h)]{\includegraphics[width=0.125\textwidth]{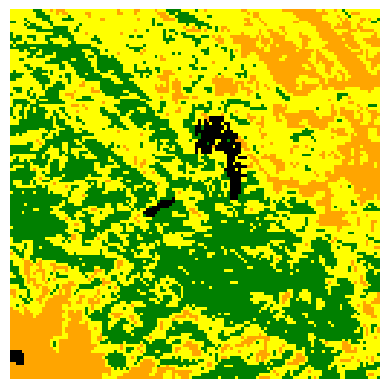}} 
\subfloat[$\alpha=0.3$ (36h)]{\includegraphics[width=0.125\textwidth]{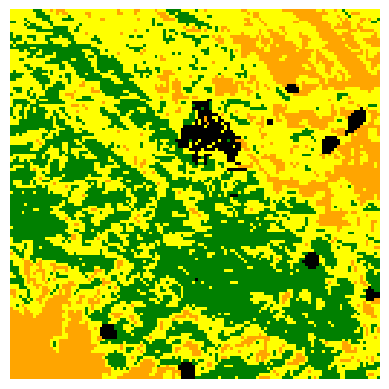}} 
\subfloat[$\alpha=0.3$ (42h)]{\includegraphics[width=0.125\textwidth]{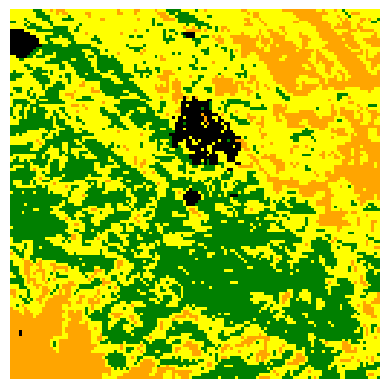}} 
\subfloat[$\alpha=0.3$ (48h)]{\includegraphics[width=0.125\textwidth]{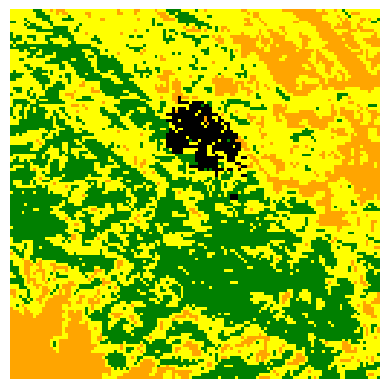}}\\
\subfloat[$\beta=1.0$ (30h)]{\includegraphics[width=0.125\textwidth]{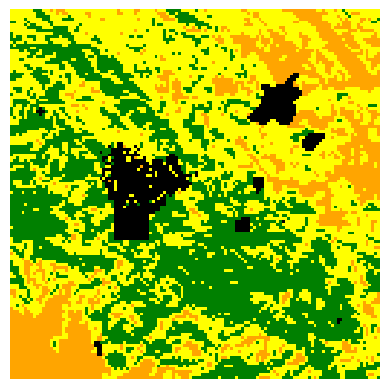}} 
\subfloat[$\beta=1.0$ (36h)]{\includegraphics[width=0.125\textwidth]{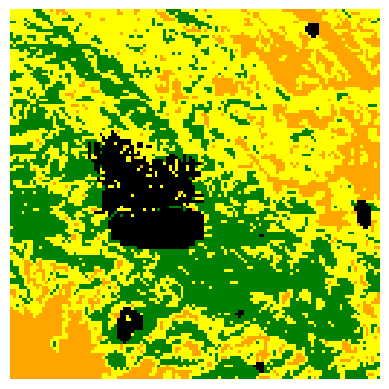}} 
\subfloat[$\beta=1.0$ (42h)]{\includegraphics[width=0.125\textwidth]{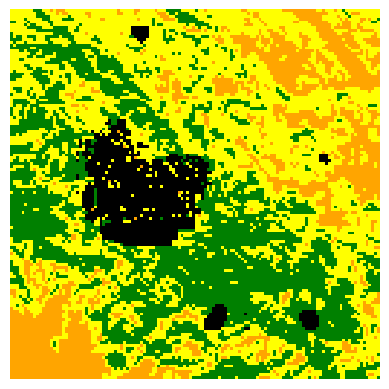}} 
\subfloat[$\beta=1.0$ (48h)]{\includegraphics[width=0.125\textwidth]{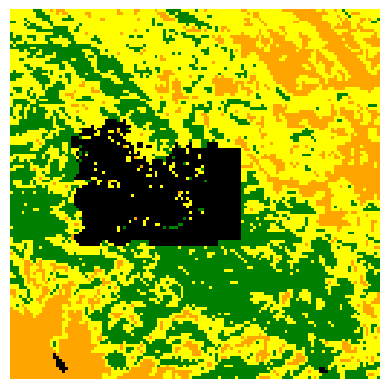}}\\
\caption{Generated wildfire events with $\alpha=0.3$ (first row) and $\beta=1.0$ (second row) for parameter studies.}
\label{fig:ablation_coefficient}
\end{figure}

\rev{The parameter $\beta$ determines the relative weight of $\mathcal{L}_{commit}$ while $\alpha$ determines the level of additional noise when generating synthetic fire spread videos. A small value of $\alpha$ will lead to a generated video that is very similar to the input due to the reconstruction loss. When the value of the parameter $\alpha$ exceeds a certain threshold, it engenders a diminution in controllability and an increase in noise levels within the generated video, as shown in Figure \ref{fig:ablation_coefficient}.}

\rev{On the other hand, the coefficient $\beta$ balances the weight of the commit loss. When the value is set excessively high, it will additionally introduce unrealistic noises in the generated scenarios, as depicted in Figure \ref{fig:ablation_coefficient}.}

\rev{These experiments illustrate the necessity of making thoughtful parameter selections during the training and application of the generative model.}

\subsection{Number of generated videos in the training set}

\begin{figure*}[h!]
\centering
\includegraphics[width=0.95\textwidth]{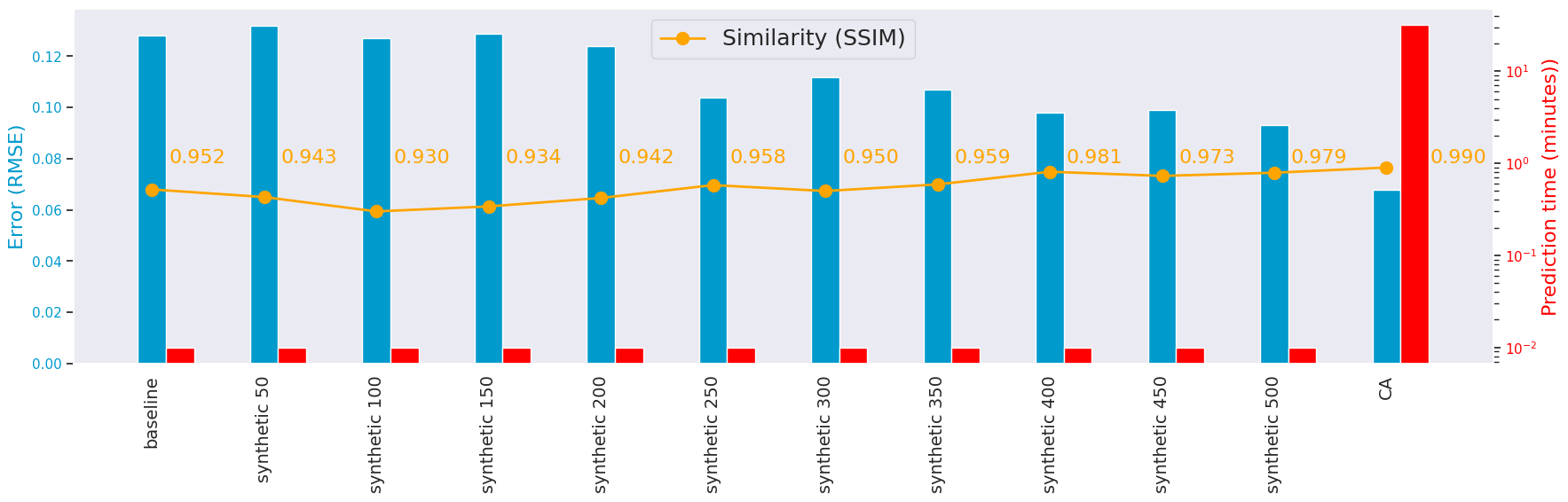}
\caption{Performance of surrogate models trained with different numbers of generated wildfire videos }
\label{fig:hist_ab}
\end{figure*}

\rev{In order to delve deeper into the influence of the number of generated fire scenarios during the training process, we conducted experiments where the surrogate model was trained using a range of 50 to 500 generated fire videos. Subsequently, we assessed the performance of the trained surrogate model, as depicted in Figure \ref{fig:hist_ab}.}

\rev{The results clearly demonstrate that the performance of the surrogate model exhibits a noticeable improvement as the number of generated videos in the training data increases. The performance stabilizes once the number of generated videos exceeds four hundred. Notably, it is important to highlight that employing a small number of generated videos may result in a deterioration of the prediction performance. }

\section*{Acknowledgements}
The authors thank Dr. Yuhang Huang and Prof. Yufang Jin from University of California, Davis for data curation of the Chimney fire burned area. The authors thank Mr. Bo Pang from Imperial College London for runing the MTT fire simulation for comparison.
 This research is funded by the Leverhulme Centre for Wildfires, Environment and Society through the Leverhulme Trust, grant number RC-2018-023. This work is partially supported by the EP/T000414/1 PREdictive Modelling with QuantIfication of UncERtainty for MultiphasE Systems (PREMIERE). 
 
\bibliographystyle{IEEEtran}
\bibliography{IEEEabrv,Bibliography}

\vfill


\end{document}